\documentclass[10pt,twocolumn,letterpaper]{article}

\usepackage{cvpr}              % To produce the CAMERA-READY version
\usepackage{dsfont}
\usepackage{multicol}
\usepackage{multirow}

\usepackage[dvipsnames]{xcolor}

\definecolor{cvprblue}{rgb}{0.21,0.49,0.74}
\usepackage[pagebackref,breaklinks,colorlinks,citecolor=cvprblue]{hyperref}

\def\paperID{2959} % *** Enter the Paper ID here
\def\confName{CVPR}
\def\confYear{2024}

\title{Towards the Uncharted: Density-Descending Feature Perturbation \\ for Semi-supervised Semantic Segmentation}

\author{Xiaoyang Wang$^{1,2,3}$ \hspace{15pt} Huihui Bai$^4$ \hspace{15pt} Limin Yu$^1$ \hspace{15pt} Yao Zhao$^4$ \hspace{15pt} Jimin Xiao$^{1}$\thanks{Corresponding author.} \\
    $^1$XJTLU\hspace{15pt} $^2$University of Liverpool\hspace{15pt} $^3$Metavisioncn\hspace{15pt} $^4$Beijing Jiaotong University\\ 
}

\begin{document}
\maketitle

\begin{abstract}
Semi-supervised semantic segmentation allows model to mine effective supervision from unlabeled data to complement label-guided training. Recent research has primarily focused on consistency regularization techniques, exploring perturbation-invariant training at both the image and feature levels. In this work, we proposed a novel feature-level consistency learning framework named Density-Descending Feature Perturbation (DDFP). Inspired by the low-density separation assumption in semi-supervised learning, our key insight is that feature density can shed a light on the most promising direction for the segmentation classifier to explore, which is the regions with lower density. We propose to shift features with confident predictions towards lower-density regions by perturbation injection. The perturbed features are then supervised by the predictions on the original features, thereby compelling the classifier to explore less dense regions to effectively regularize the decision boundary. Central to our method is the estimation of feature density. To this end, we introduce a lightweight density estimator based on normalizing flow, allowing for efficient capture of the feature density distribution in an online manner. By extracting gradients from the density estimator, we can determine the direction towards less dense regions for each feature. The proposed DDFP outperforms other designs on feature-level perturbations and shows state of the art performances on both Pascal VOC and Cityscapes dataset under various partition protocols. The project is available at \url{https://github.com/Gavinwxy/DDFP}.
\end{abstract}

\section{Introduction}
Semantic segmentation is a fundamental task in visual understanding, involving pixel-level classification on input images~\cite{fcn,deeplab,Zhao2017PyramidSP}. However, segmentation models often exhibit a strong dependence on large amounts of annotated data. Unfortunately, collecting such training data can be both time-consuming and laborious, thereby impeding the practical application of segmentation models. To address the challenge, semi-supervised semantic segmentation is drawing growing attention recently~\cite{Souly2017SemiSS}. Such learning paradigm aims to enhance label-efficiency by utilizing a limit amount of labeled data alongside massive unlabeled data. The key lies in mining effective training signal from unlabeled data to allow better model generalization.

\begin{figure}
    \centering
    \includegraphics[width=\columnwidth]{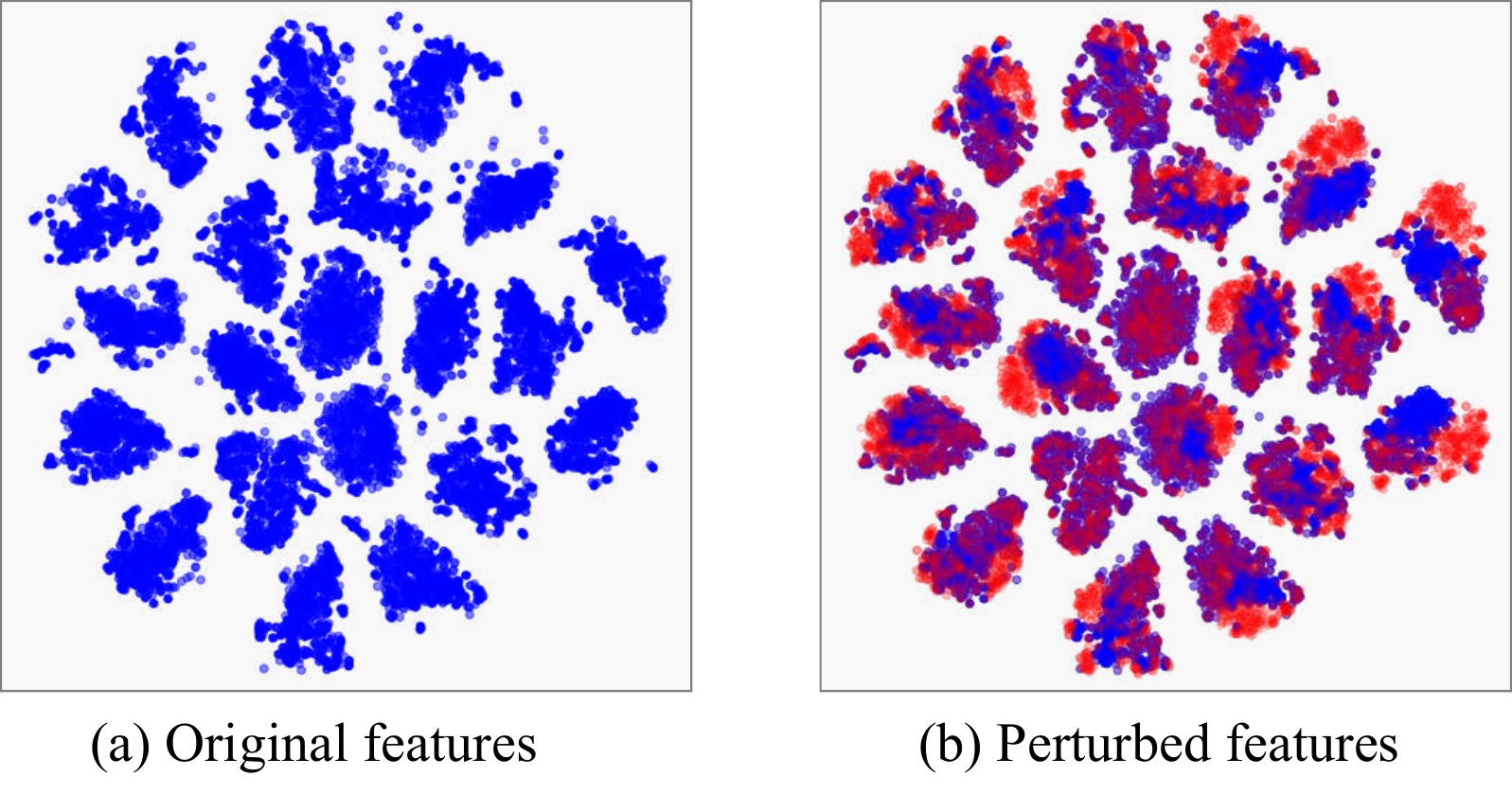}
    \caption{t-SNE visualization of per-pixel features from Pascal VOC 2012 dataset~\cite{pascal}. (a) Features extracted from encoder. (b) Features after the proposed DDFP strategy (shown in red). The perturbed features significantly deviate from high density centers and move towards low density regions within and out of clusters.}
    \label{fig:motivation}
\end{figure}

Recent research in semi-supervised semantic segmentation has witnessed a transition from primal adversarial learning approaches~\cite{Souly2017SemiSS,Hung2018AdversarialLF,Mittal2021SemiSupervisedSS} to self-training methods~\cite{Yuan2021ASB,He2021RedistributingBP,Yang2022STMS,Guan2022UnbiasedSR}. Recent studies focus on consistency regularization frameworks, which aim to enforce prediction agreement across diverse views of unlabelled images~\cite{DBLP:conf/bmvc/FrenchLAMF20}. Notably, alternative data views can also be created at feature level, which is explored by a line of works~\cite{Ouali2020SemiSupervisedSS,Liu2022PerturbedAS,Yang2022RevisitingWC}. Among these methods, uniformly sampled noise, random channel dropout, and perturbations adversarial to predictions are applied to image features. Subsequently, predictions on perturbed features are supervised by those derived from the original ones, enabling feature-level consistency learning. These methods have demonstrated their efficacy. Nonetheless, previous design 
of feature-level perturbations seems to be designed for general purpose but not tailored for the context of semi-supervised learning.

% FixMatch~\cite{Sohn2020FixMatchSS} proposes to generate pseudo labels from weakly augmented unlabeled images to supervise their strongly augmented counterparts, thereby inspiring subsequent works~\cite{,Chen2021SemiSupervisedSS,Hu2021SemiSupervisedSS,Huo2021ATSOAT}.
% Nonetheless, previous designs 
% on feature-level perturbations are not tailored for semi-supervised learning and appear to lack a clear basis.

The low-density separation assumption in semi-supervised learning states that decision boundaries should ideally reside in low-density regions within the feature space~\cite{Chapelle2005SemiSupervisedCB}. While previous efforts implicitly move towards this objective, the question is whether a more direct and effective approach can be devised. In an effort to address this question, we propose a novel feature perturbation strategy called Density-Descending Feature Perturbation (DDFP) for semi-supervised semantic segmentation. We assume that density information in feature space can shed a light on the direction to improve decision boundary. With density information, features with reliable label guidance in self-training as shown in Fig.~\ref{fig:motivation} (a) can be perturbed toward regions of lower density as in Fig.~\ref{fig:motivation} (b), while still be supervised by its original semantic. Hence, the decision boundary will be forced to explore less dense regions in feature space to prevent classifier overfitting easy patterns.   

The crux in our method is the acquisition of feature density distribution. Normalizing flow~\cite{Dinh2016DensityEU, Kingma2018GlowGF}, designed for generative modelling, is a perfect fit for this task. Hence, we propose a density estimator based on a normalizing flow to learn and predict feature density in real time upon the training of segmentation model. The estimator constructs bijective mappings transforming a predefined base distribution into the target feature probability density, with the mappings optimized by likelihood maximization. Inspired by previous work~\cite{Izmailov2019SemiSupervisedLW}, we initialize the base distribution as a Gaussian mixture model, where pair-wise links are built between each Gaussian component and semantic category, enabling more fine-grained optimization and density description. Once the density information is obtained, the density-descending direction on features can be obtained from the density objective as the gradient over original features. Hence, density-descending features can be created with such perturbation injected, which are then leveraged in consistency regularization framework. The density estimator solely act as an observer upon the main segmentation network, online tracking the feature density but not directly contribute to the training of the main model. In inference, the estimator is discarded thereby avoiding any computational overhead. The knowledge learned on the feature distribution indirectly benefits the segmentation classifier, providing effective hints for its optimization.

To verify the effectiveness of the proposed method, we evaluate our method on mainstream benchmarks Pascal VOC 2012~\cite{pascal} and Cityscapes~\cite{cityscapes} dataset under different data partition settings, where our method achieves state-of-the-art performance. The contributions of our method can be summarised as following:
\begin{itemize}
    \item Inspired by the low-density separation assumption, we propose to utilize density information in feature space and design a novel density-descending feature-level perturbations for consistency regularization framework.
    \item We propose to leverage a normalizing-flow-based density estimator to online capture feature density through likelihood maximization training, from which density-descending directions can be obtained.
    \item The proposed feature-level consistency regularization achieves competitive performance on mainstream benchmark for semi-supervised semantic segmentation. 
\end{itemize}

\section{Related Works}
\begin{figure*}
    \centering
    \includegraphics[width=\textwidth]{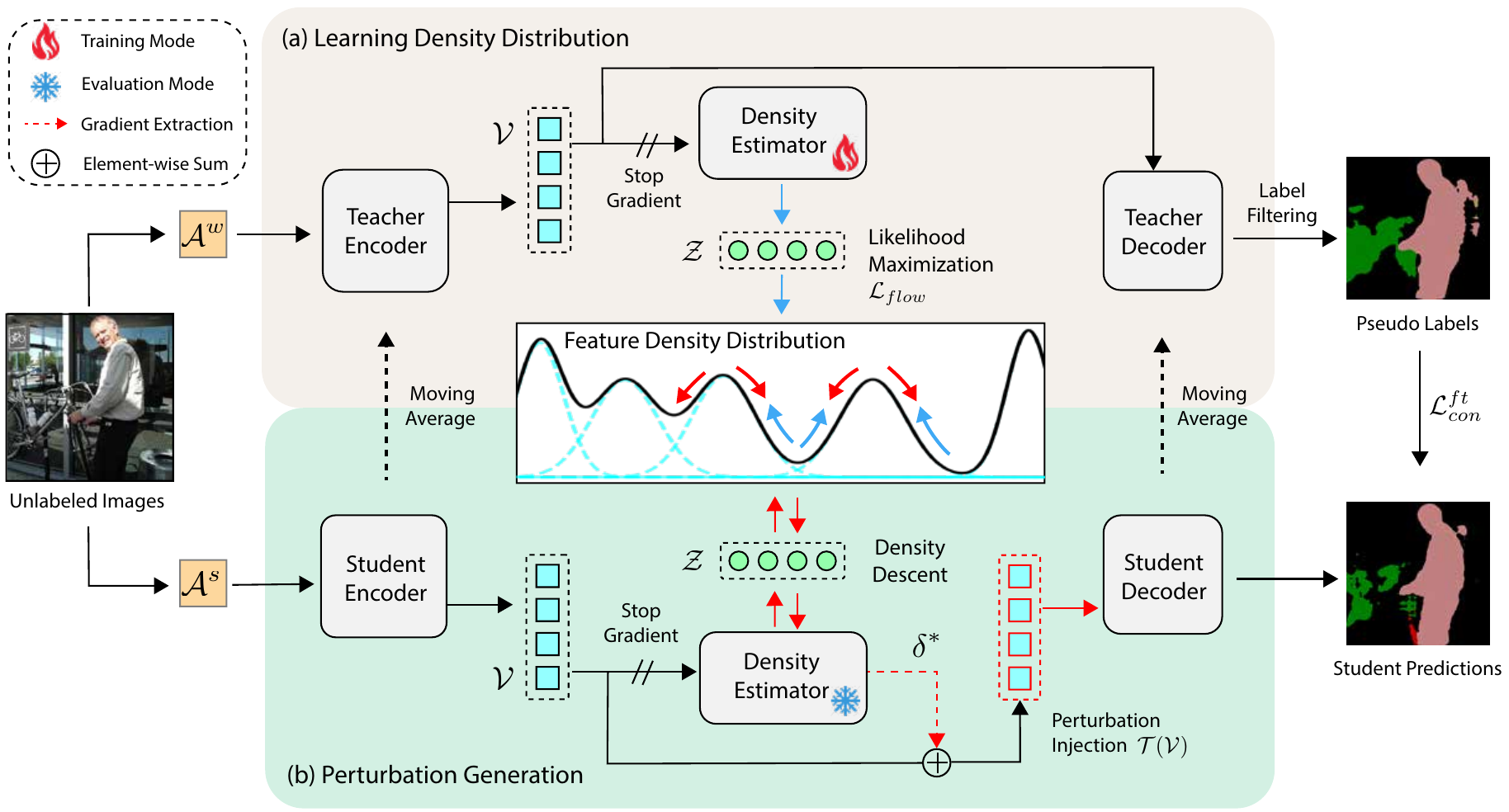}
    \caption{Overview of the proposed density-descending feature perturbation strategy. Based on the weak-to-strong consistency regularization, our method consists of two main components: (a) Learning Density Distribution and (b) Perturbation Generation. In phase (a), a lightweight normalizing-flow-based estimator is adopted to learn the density information on unlabeled features from teacher encoder. A mixture of Gaussian distribution is initialized and the estimator is optimized to maximize the feature likelihood on that distribution, which is denoted by blue arrows. Meanwhile in phase (b), the density estimator is set as evaluation mode and applied on the student features. Once feature distribution is approximated, the density-descending direction can be extracted by the gradient of the density objective on unlabeled features. Predictions on the density-descending features are supervised by pseudo labels from teacher model. The whole perturbation generation and injection process is indicated by red arrows.}
    \label{fig:main framework}
\end{figure*}

\noindent
\textbf{Semi-supervised Learning.} Semi-supervised learning (SSL) aims to mine effective supervision from unlabeled data. A fundamental technique is self-training~\cite{Rosenberg2005SemiSupervisedSO,Xie2020SelfTrainingWN,Zoph2020RethinkingPA,Pham2020MetaPL}, which generates pseudo labels for unlabeled data based on the knowledge from labeled samples, followed by re-training on the combined data to improve model generalization. Recent research focuses on consistency regularization~\cite{Laine2017TemporalEF,Sajjadi2016RegularizationWS,Tarvainen2017MeanTA,Jeong2019ConsistencybasedSL,Berthelot2020ReMixMatchSL,9656684,Liu2023ProgressiveSM}, where models are benefited from perturbation-invariant training on unlabeled samples. Among these approaches, MixMatch~\cite{Berthelot2019MixMatchAH} introduces label-guessing by averaging predictions on multiple augmented versions of unlabeled data. FixMatch~\cite{Sohn2020FixMatchSS} employs a weak-to-strong consistency strategy where pseudo labels are generated from weakly augmented samples and used to supervise strongly augmented counterparts. Subsequent works have proposed sophisticated pseudo label filtering strategies. Among them, FlexMatch~\cite{Zhang2021FlexMatchBS} takes into account the varying learning difficulties among categories and designs class-specific thresholds. FreeMatch~\cite{Wang2022FreeMatchST} proposes an adaptive threshold that adjusts based on the model's training status, while SoftMatch~\cite{Chen2023SoftMatchAT} introduces a truncated Gaussian function as confidence threshold for unlabeled samples.

\vspace{2mm}
\noindent
\textbf{Semi-supervised Semantic Segmentation.}
Research in semi-supervised semantic segmentation has been influenced by advancements in SSL techniques. Notably, self-training and co-training methods~\cite{Feng2020DMTDM,Yang2022STMS,Chen2021SemiSupervisedSS,Wang2023ConflictBasedCC}  have demonstrated success by extracting pseudo labels from either a single model or multiple models. For instance, ST++~\cite{Yang2022STMS} progressively and selectively generates pseudo labels to ensure high-quality re-training. CPS~\cite{Chen2021SemiSupervisedSS} utilizes two differently initialized models and exchanges pseudo labels between them to facilitate cross-supervision, while CCVC~\cite{Wang2023ConflictBasedCC} follows a similar framework but incorporates discrepancy loss to enhance model diversity. Another research line explores the utilization of contrastive learning~\cite{Alonso2021SemiSupervisedSS, Zhong2021PixelCS,Lai2021SemisupervisedSS,Zhao2021ContrastiveLF,Zhou2021C3SemiSegCS,Wang2022SemiSupervisedSS,liu2022reco,Wang2023HuntingSD}. Among these methods, ReCo~\cite{liu2022reco} performs contrastive learning on hard negative samples to regularize the feature space, and U2PL~\cite{Wang2022SemiSupervisedSS} extracts negative samples from unreliable predictions to contrast against positive samples.

Consistency regularization has also shown progress in semi-supervised semantic segmentation~\cite{DBLP:conf/bmvc/FrenchLAMF20,Hu2021SemiSupervisedSS,Yang2022RevisitingWC,Ouali2020SemiSupervisedSS,Liu2022PerturbedAS,Zhao2022AugmentationMA,Zhao2024SFCSF}. French et al.~\cite{DBLP:conf/bmvc/FrenchLAMF20} adapts CutMix and CutOut techniques from image classification to the segmentation domain, serving as a baseline for image-level strong augmentations. Subsequently, AEL~\cite{Hu2021SemiSupervisedSS} proposes adaptive CutMix, which targets under-performing categories during training. Consistency has also been explored at the feature level. UniMatch~\cite{Yang2022RevisitingWC} introduces random channel dropout on features and enforces consistency between predictions on perturbed features and the original ones. CCT~\cite{Ouali2020SemiSupervisedSS} creates a pool of feature perturbation strategies, including random noise, spatial dropout, and adversarial perturbations, which are applied to multiple auxiliary decoders. PS-MT~\cite{Liu2022PerturbedAS} further explores adversarial perturbations that induce the most disagreement among multiple teacher models, effectively regularizing the training of the student model.

Our research also focuses on feature-level consistency regularization. However, our novelty lies in the design of feature-level perturbations that leverage density information in the feature space to regulate the decision boundary.

\section{Methodology}

\subsection{Problem Statement}
In semi-supervised semantic segmentation, an labeled set $\mathcal{D}^{l} = \{(x_i^l, y_i^l)\}_{i=1}^{|\mathcal{D}^{l}|}$ is given with images $x_i^l \in \mathbb{R}^{H\times W\times C}$ of size $H \times W$ and channel number $C$. $y_i^l \in \mathbb{R}^{H\times W \times K}$ are annotations with $K$ classes. Meanwhile, massive unlabeled data $\mathcal{D}^u = \{x_i^u\}_{i=1}^{|\mathcal{D}^u|}$ are also provided, where $|\mathcal{D}^u| \gg |\mathcal{D}^l|$. Then, semi-supervised segmentation models are designed to be optimized on both data sets, aiming to achieve stronger model generalization beyond labeled data.

\subsection{Basic Framework}
Before diving into the proposed feature perturbation strategy, we first go through the basic framework. We adopt the widely used teacher-student model combined with image-level consistency regularization in this work. The segmentation model $f = g \circ h$ consists of a feature encoder $h(\cdot): \mathcal{X} \rightarrow \mathcal{V}$, which maps images $\mathcal{X}$ into feature space $\mathcal{V}$, and a mask decoder $g(\cdot): \mathcal{V} \rightarrow \mathcal{P}$, which then decodes features into class probabilities $\mathcal{P}$. While the student model is optimized, the teacher model $f' = g' \circ h'$ is updated as the exponential moving average (EMA) of student. 

For the supervised learning part, a mini-batch of labeled images $\mathcal{B}^l = \{(x^l_i, y^l_i)\}_{i=1}^{|\mathcal{B}^l|}$ is given. With $i$ for $i$-th image and $j$ indicating pixel index, the cross-entropy loss is applied to supervise the model predictions:
\begin{equation}
    \mathcal{L}_{sup} = \frac{1}{|\mathcal{B}^l|} \frac{1}{HW} \sum_{i=1}^{|\mathcal{B}^l|} \sum_{j=1}^{HW}\ell_{ce} (f(x^l_{ij}), y^l_{ij}).
\end{equation}

The image-level consistency regularization is performed on unlabeled images $\mathcal{B}^u = \{x^u_i\}_{i=1}^{|\mathcal{B}^u|}$. Let $\mathcal{A}^w(\cdot)$ and $\mathcal{A}^s(\cdot)$ denote the weak and strong image augmentation strategies, respectively. For an unlabeled image $x_{i}^u$, we obtain the probability predictions on its augmented versions as: 
\begin{align}
    p_{i}^w &= f'(\mathcal{A}^w(x_{i}^u)) \\
    p_{i}^s &= f(\mathcal{A}^s(x_{i}^u)).
\end{align}
Pseudo labels $y_{ij}^u$ are extracted by one-hot encoding on teacher predictions $p_{ij}^w$ and then used to supervised the student predictions $p_{ij}^s$. The image-level consistency loss $\mathcal{L}_{con}^{im}$ is calculated as:
\begin{equation}
\mathcal{L}_{con}^{im} = \frac{1}{|\mathcal{B}^u|} \frac{1}{HW} \sum_{i=1}^{|\mathcal{B}^u|} \sum_{j=1}^{HW}\ell_{ce}(p_{ij}^s, y^u_{ij})\cdot \mathds{1} (\max(p_{ij}^w) > \tau),
\label{eq:image-level consistency}
\end{equation}
where a fixed probability threshold $\tau$ is applied to screen out potential noisy labels with low prediction confidence.

\subsection{Density-Descending Feature Exploration}
Upon the basic learning framework, we introduce a novel feature-level consistency regularization to boost the model generalization ability. This section gives details of our density-descending feature perturbation strategy. We will first introduce the proposed density estimator which is based on normalizing flow. Then, the whole process is explained as shown in Fig.~\ref{fig:main framework}, which mainly consists two main sessions: (a) Learning Density Distribution where density estimator is optimized by likelihood maximization on image features to capture their density, and (b) Perturbation Generation where the learned density information is leveraged to generate density-descending perturbations on features.

\subsubsection{Feature Density Estimation Module}

To address the challenge of tracking the unknown density distribution $p_{\mathcal{V}}$ in a dynamic feature space $\mathcal{V}$, we propose a density estimator based on normalizing flows. Normalizing flows are specifically designed for generative modeling tasks, allowing for the learning of complex density functions by transforming a known distribution through a series of invertible mappings. In our approach, given a known latent distribution $p_{\mathcal{Z}}$, the density estimator is defined as the mapping $\varphi(v)$ that transforms features $\mathcal{V} \in \mathbb{R}^{d}$ into the latent space $\mathcal{Z} \in \mathbb{R}^{d}$, with the inverse mapping $\varphi^{-1}(z)$ to map them back to the original feature space.

The density of the unknown distribution $p_{\mathcal{V}}(v)$ can be modeled using the transformed variable $\varphi(v)$ through the change of variable formula:
\begin{equation}
p_{\mathcal{V}}(v) = p_{\mathcal{Z}}(\varphi(v)) \cdot \left|\det\frac{\partial \varphi}{\partial v} \right|,
\end{equation}
where $\frac{\partial \varphi}{\partial v}$ represents the Jacobian matrix of the transformations in $\varphi$, which are carefully designed to ensure that the Jacobian determinant is tractable. The parameters in the normalizing flow are optimized through likelihood maximization on $p_{\mathcal{V}}$.

\subsubsection{Learning Feature Density}

During the training of segmentation model, the pixel-level features on both labeled and unlabeled images are extracted by encoders. Since teacher encoder shows higher stability in training, we focus on optimizing the density estimator $\varphi_{\theta}$, parameterised by $\theta$, on teacher features. In the latent space, we assign distinct Gaussian distributions to each class. Specifically, the distribution for class $k$ is initialized with mean $\mu_{k}$ and covariance $\Sigma_{k}$, while the overall distribution across all categories is modeled as a mixture of Gaussians, weighted by $\pi_{k}$ where $\sum_{k=1}^{K} \pi_{k} = 1$. This allows us to estimate the likelihood of a latent variable $z$ with an unknown label using the following expression:   
\begin{equation}
    p_{\mathcal{Z}}(z) =  \sum_{k=1}^{K} \pi_{k} \mathcal{N}(z | \mu_k, \Sigma_k).
\end{equation}

Specifically, in each training iteration, a batch of labeled features $\mathcal{V}^l = \{v_m^l, y_m\}_{m=1}^{|\mathcal{V}^l|}$ and unlabeled features $\mathcal{V}^u = \{v_n^u\}_{n=1}^{|\mathcal{V}^u|}$ are collected from teacher encoder with gradient cut off. For a labeled feature $v_{m}^{l}$ of class $k$, its likelihood is estimated by the target Gaussian component as
\begin{equation}
    p_{\mathcal{V}}(v_{m}^{l} | y=k; \theta) =  \mathcal{N}(\varphi_{\theta}(v_{m}^{l}) | \mu_k, \Sigma_{k}) \cdot \left|\det \frac{\partial \varphi_{\theta}}{\partial v_{m}^{l}}\right|.
    \label{eq:sup likelihood}
\end{equation}

For unlabeled feature, since no trustworthy labels are provided, their density can be captured in unsupervised manner to alleviate the potential bias in classifier predictions:
\begin{equation}
    p_{\mathcal{V}}(v_{n}^{u}; \theta) = \sum_{k=1}^{K} \pi_{k} \mathcal{N}(\varphi_{\theta}(v_{n}^{u}) | \mu_{k}, \Sigma_{k}) \cdot \left|\det \frac{\partial \varphi_{\theta}}{\partial v_{n}^{u}} \right|.
    \label{eq:unsup likelihood}
\end{equation}

The optimization objective $\mathcal{L}_{flow}$ for density estimator is the unified log-likelihood combining estimation from both labeled and unlabeled features:
\begin{equation}
\begin{split}
\mathcal{L}_{flow} = - \frac{1}{|\mathcal{V}^{l}|+|\mathcal{V}^u|}\Biggl( \sum_{m=1}^{|\mathcal{V}^{l}|} \log p_{\mathcal{V}}(v^{l}_{m} | y_m; \theta) \\
+ \sum_{n=1}^{|\mathcal{V}^u|} \log p_{\mathcal{V}}(v^{u}_{n}; \theta) \Biggl). 
\end{split}
\end{equation}
Then, $\theta$ is optimized to maximize likelihood in feature space. By learning upon $p_{\mathcal{Z}}$, we can effectively harness the knowledge present in the labeled set to establish connections between classes and their corresponding Gaussian components and to guide the initialization of the density estimator $\varphi_{\theta}$. This approach allow us to capture the inherent structure and relationships between different classes. Additionally, the optimization process on the unlabeled features through unsupervised likelihood maximization enables the density estimator to adapt and refine its representation to better align with the underlying data distribution.

\subsubsection{Generating Density-descending Perturbation}
During the optimization on the density estimator $\varphi_{\theta}$, we employ it to generate density-descending perturbations on student features for consistency learning. The log-likelihood evaluation of a student feature $v$ in the Gaussian mixtures, denoted as $\log p_{\mathcal{V}}(v; \theta)$, is determined based on the current $\varphi_{\theta}$ using Eq.~\ref{eq:unsup likelihood}. In such setting, each feature is evaluated in the unified distribution rather than a single Gaussian component, assuring the following density descent happens in the global scale.

The generation of density-descending perturbations draws inspiration from adversarial learning practices. However, unlike traditional adversarial learning, where perturbations primarily aims to attack classification results, our strategy focuses on the density objective, guided by the evaluation of the estimator $\varphi_{\theta}$. Within a predefined exploration range $\epsilon$, the objective is to find perturbations $\delta^*$ that result in the most substantial decrease in feature density:
\begin{equation}
\delta^* = \underset{{\lVert \delta \rVert_{2} \leq \epsilon}}{\operatorname{argmax}} \left(-\log p_{\mathcal{V}}(v + \delta)\right).
\label{eq:definition on pt}
\end{equation}

The direction of perturbation can be determined by the gradient of the likelihood minimization objective $-\log p_{\mathcal{V}}(v)$ over the target feature $v$, denoted as $\nabla_{v} (-\log p_{\mathcal{V}}(v))$. The calculation is as follows:
\begin{equation}
\delta^* = \epsilon \cdot \lVert \nabla_{v} (-\log p_{\mathcal{V}}(v)) \rVert_{2},
\label{eq:perturbation calculation}
\end{equation}
where $\lVert \cdot \rVert_{2}$ represents L2-normalization, and $\epsilon$ the magnitude of the perturbation, determining the exploration step. Consequently, the density-descending version of the original feature can be obtained by injecting the perturbation:
\begin{equation}
\mathcal{T}(v) = v + \delta^*,
\end{equation}
where $\mathcal{T}(\cdot)$ denotes the perturbation injection operation. The perturbed features are expected to shift towards lower-density regions, based on the current estimation of $\varphi_{\theta}$.

\subsection{Unified Training Objective} 
With the proposed perturbation generation strategy, we employ feature-level consistency regularization. The features to be perturbed are shared with image-level consistency learning for computational efficiency, which are extracted from the student encoder $h$ as follows:
\begin{equation}
v_{i}^s = h(\mathcal{A}^s(x_{i}^u)).
\end{equation}
Subsequently, predictions on these features are generated using the student decoder $g$:
\begin{equation}
p_{ij}^{ft} = g(\mathcal{T}(v_{ij}^s)).
\end{equation}
To align the predictions between the original and perturbed features, we introduce the consistency loss $\mathcal{L}_{con}^{ft}$, which is calculated as:
\begin{equation}
    \mathcal{L}_{con}^{ft} = \frac{1}{|\mathcal{B}^u|} \frac{1}{HW} \sum_{i=1}^{|\mathcal{B}^u|} \sum_{j=1}^{HW}\ell_{ce}(p_{ij}^{ft}, y^u_{ij})\cdot \mathds{1} (\max(p_{ij}^w) > \tau).
    \label{eq:feature-level consistency}
\end{equation}
Here, pseudo labels are derived from teacher predictions, and label filtration is employed to ensure that density-descending features are only guided by reliable labels.

The overall optimization is achieved by unifying three learning objectives: the supervised learning loss $\mathcal{L}_{sup}$, the image-level consistency loss $\mathcal{L}_{con}^{im}$, and the feature-level consistency loss with density-descending feature perturbations. The unified objective $\mathcal{L}_{uni}$ is defined as:
\begin{equation}
\mathcal{L}_{uni} = \mathcal{L}_{sup} + \mathcal{L}_{con}^{im} + \lambda_{ft}\mathcal{L}_{con}^{ft}.
\label{eq:overall optimization}
\end{equation}
Here, $\lambda_{ft}$ represents the weight assigned to the feature-level consistency loss term.

\section{Experiments}
\subsection{Implementation Details.}

\vspace{2mm}
\noindent
\textbf{Datasets.} The experiments are performed on two standard datasets, which are Pascal VOC 2012~\cite{pascal} and Cityscapes~\cite{cityscapes}. Pascal VOC contains 20 foreground and 1 background classes. It is initially built with 1464 training images with high-quality annotations and 1449 validation images, which is denoted as \textit{classic} set. Then, SBD dataset~\cite{sbd} with coarse annotations is introduced to extend the training set to 10582 images to form the \textit{blended} set. We conduct experiments on both \textit{classic} and \textit{blended} sets. Cityscapes contains 19 semantic categories of urban scenes. It consists of 2975 and 500 annotated images for training and validation, respectively. For both datasets, we follow data partitions in CPS~\cite{Chen2021SemiSupervisedSS} to generate subsets of 1/16, 1/8, 1/4, and 1/2 from the training set as labeled data, while the remaining images are utilized as the unlabeled set. 

\vspace{2mm}
\noindent
\textbf{Evaluation Protocols.} We report the mean of intersection over union (mIoU) for all settings. For Pascal VOC, single-scale evaluation on center-cropped images are performed. For Cityscapes, we adopt sliding window evaluation on the validation images with resolution of $1024 \times 2048$.

\vspace{2mm}
\noindent
\textbf{Implementation Details.} We adopt DeepLabV3+~\cite{Chen2018EncoderDecoderWA} with ResNet-101~\cite{He2016DeepRL} pretrained on ImageNet~\cite{Deng2009ImageNetAL} as the segmentation model. We use SGD optimizer with momentum of 0.9 and polynomial learning rate scheduler. For Pascal VOC, the model is trained for 80 epochs with initial learning rate of 0.001. The images are copped to 513$\times$ 513 for training. For Cityscapes, we train 200 epochs with initial learning rate of 0.01 under a crop size of 769$\times$769. The batch size is set to 16 for both datasets. The prediction thresholds $\tau$ in Eq.~\ref{eq:image-level consistency} and \ref{eq:feature-level consistency} are set to 0.95 for Pascal VOC and 0.7 for Cityscapes. The momentum for updating teacher model is set to 0.999 for all experiments. 

For the weak image-level augmentation $\mathcal{A}^{w}(\cdot)$, we adopt random resize within the scale range $[0.5, 2.0]$, random crop and random flip. The strong data augmentation $\mathcal{A}^{s}(\cdot)$ is implemented by the random combination of CutMix~\cite{DBLP:conf/bmvc/FrenchLAMF20}, Gaussian blur, color jitter and random grayscale.

\begin{figure}
    \centering
    \includegraphics[width=\linewidth]{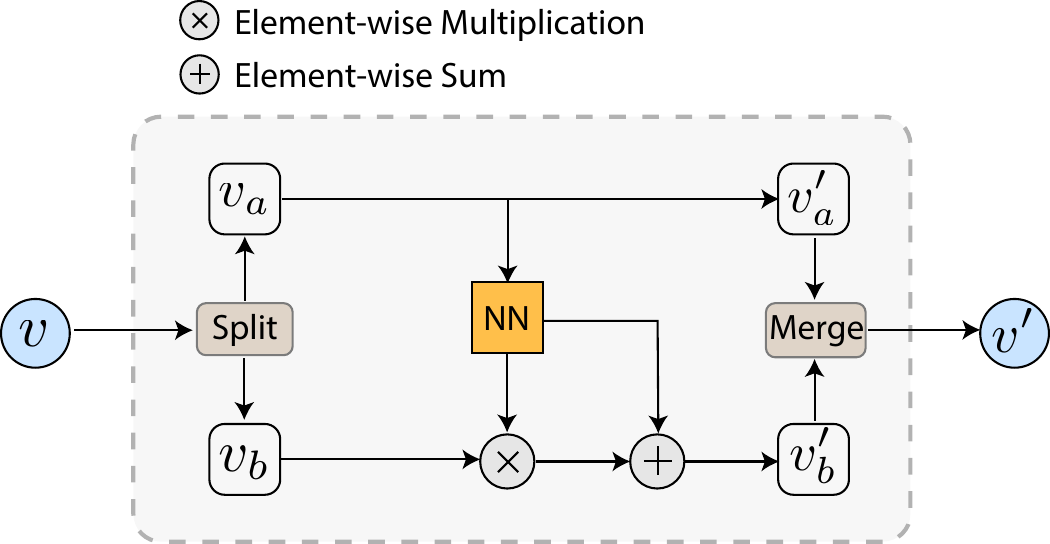}
    \caption{Block design for the proposed density estimator.}
    \label{fig:flow module}
\end{figure}

\vspace{2mm}
\noindent
\textbf{Density Estimator.} Our density estimator is designed as a modified RealNVP~\cite{Dinh2016DensityEU} module. Since the features are regulated by the segmentation objective, they show more clear patterns and also with lower dimensions compared to natural images, which largely ease the process of density learning. This allows a highly lightweight density learner with negligible amount of parameters. Fig.~\ref{fig:flow module} shows a single block in the proposed density estimator, where the input feature $v$ is split in half in channel and then merged into $v'$ with identical dimensions. The whole module contains two cascaded blocks with channel permutation between them, transforming from $v$ to $z$. The NN part is implemented by two cascaded Linear (256) layers with learnable parameters. The module is implemented with FrEIA~\cite{freia} library.  

For the GMM in latent space, the number of Gaussian components is set as 21 and 19 for Pascal VOC and Cityscapes, respectively. The mean for $k$-th component is drawn from the standard normal distribution $\mu_{k} \sim \mathcal{N}(0,I)$ and the covariance matrix is set as an identity matrix $\Sigma_{k}=I$. Only the flow module $\varphi_{\theta}$ is updated by likelihood maximization. In each iteration, 20k feature vectors are sampled equally from both labeled and unlabeled samples for training. For all the experiments, we adopt Adam optimizer and step learning rate scheduler with initial learning rate of 0.001. The training of density estimator starts from the second epoch during the training of segmentation model. 

% We found that training with sampled data yields satisfying density estimation results. To track the training status of the density estimator, we treat it as a classifier on features and evaluate its pixel-level classification performance.

% Since the quality of density estimation cannot be directly evaluated, we leverage the density estimator as a classifier where features are classified based on the predicted likelihood in each Gaussian components. The estimator is tracked by evaluating its pixel-level classification performance. We adopt Adam optimizer with learning rate of 0.001.  

\subsection{Comparison with State-of-the-Art Methods}
In this section, we compare our method with the state-of-the-art on both Pascal VOC and Cityscapes under various partition protocols. The data splits in our experiments strictly follow previous works~\cite{Chen2021SemiSupervisedSS, Yang2022STMS,Liu2022PerturbedAS}. All the results are produced by DeepLabV3+ segmentation decoder with ResNet-101 as backbone.

\vspace{2mm}
\noindent
\textbf{Results on Pascal VOC 2012.} In Tab.~\ref{tab:pascal classic}, we compare our DDFP with other methods on \emph{classic} Pascal VOC set and our method shows competitive results. DDFP brings significant performance improvement over supervised baseline by +29.18\%, +23.09\% and +13.63\% on 1/16, 1/8 and 1/4 splits. Compared with previous method PS-MT~\cite{Liu2022PerturbedAS} that focuses on designing feature-level perturbations, our DDFP yields stronger performance especially in low-data regime. Specifically, we improve over PS-MT by +9.15\% and +8.43\% under 92 and 183 labeled images, respectively.

Tab.~\ref{tab:pascal blended} reports the results on \emph{blended} Pascal VOC dataset that contains noisy annotations, which is a more challenging setting. Our DDFP consistently produce competitive results. Compared with supervised baseline, our method achieves performance gain of +10.45\%, +7.33\%, +4.03\% and +3.77\% on 1/16, 1/8, 1/4 and 1/2 labeled image proportions, respectively. Compared with previous methods, our method yields best results in most cases.

\vspace{2mm}
\noindent
\textbf{Results on Cityscapes.} In Tab.~\ref{tab:Cityscapes}, we evaluate our method on Cityscapes. The proposed DDFP significantly improves supervised baseline by +11.36\%, +5.66\%, +5.45\% and +2.99\% on 1/16, 1/8, 1/4 and 1/2 splits. Our method outperforms previous best method UniMatch~\cite{Yang2022RevisitingWC} by +0.68\% and +1.32\% on 1/4 and 1/2 data settings.

\begin{table*}[t]
\centering
\caption{Comparison with state-of-the-art methods on PASCAL VOC 2012 validation set with mIoU results (\%) $\uparrow$. Labeled images are sampled from the high-quality \emph{classic} set comprising 1464 samples. The fractions $1/n$ and the following integers $(m)$ denote the proportions and numbers of labeled images, respectively.}
\resizebox{.55\textwidth}{!}{
\begin{tabular}{@{}l|ccccc@{}}
\toprule
Method          & 1/16 (92) & 1/8 (183) & 1/4 (366) & 1/2 (732) & Full (1464) \\ \midrule
Supervised & 45.77     & 54.92     & 65.88     & 71.69     & 72.50       \\ \midrule
CutMix~\cite{DBLP:conf/bmvc/FrenchLAMF20}          & 52.16     & 63.47     & 69.46     & 73.73     & 76.54       \\
% PseudoSeg~\cite{Zou2021PseudoSegDP}       & 57.60     & 65.50     & 69.14     & 72.41     & 73.23       \\
% PC$^{2}$Seg~\cite{Zhong2021PixelCS}          & 57.00     & 66.28     & 69.78     & 73.05     & 74.15       \\
CPS~\cite{Chen2021SemiSupervisedSS}             & 64.07     & 67.42     & 71.71     & 75.88     & -           \\
U$^{2}$PL~\cite{Wang2022SemiSupervisedSS}            & 67.98     & 69.15     & 73.66     & 76.16     & 79.49       \\
ST++~\cite{Yang2022STMS}            & 65.20    & 71.00   & 74.60  & 77.30  &79.10          \\
PS-MT~\cite{Liu2022PerturbedAS}           & 65.80     & 69.58     & 76.57     & 78.42     & 80.01       \\
PCR~\cite{Xu2022SemisupervisedSS} &70.06 &74.71 &77.16 &78.49 &80.65 \\
GTA-Seg~\cite{Jin2023SemiSupervisedSS} &70.02 &73.16 &75.57 &78.37 &80.47 \\
UniMatch~\cite{Yang2022RevisitingWC} &\textbf{75.20} &77.20 &78.80 &79.90 &81.20 \\
CCVC~\cite{Wang2023ConflictBasedCC}   &70.20 &74.40 &77.40 &79.10 &80.50 \\
AugSeg~\cite{Zhao2022AugmentationMA} &71.09 &75.45 &78.80 &80.33 &81.36\\
\midrule
\textbf{Ours} &74.95    &\textbf{78.01}  &\textbf{79.51}  &\textbf{81.21} &\textbf{81.96} \\
\bottomrule
\end{tabular}
}
\label{tab:pascal classic}
\end{table*}

\begin{table}[t]
\centering
\caption{Comparison with state-of-the-art methods on PASCAL VOC 2012 validation set with mIoU results (\%) $\uparrow$. Labeled images are sampled from the extended \emph{blended} set which consists of 10582 samples. * means reproduced results on CPS~\cite{Chen2021SemiSupervisedSS} splits.}
\resizebox{\columnwidth}{!}{
\begin{tabular}{@{}l|cccc@{}}
\toprule
Method          & 1/16 (662)    & 1/8 (1323)     & 1/4 (2646)     & 1/2 (5291)     \\ \midrule
Supervised & 67.87                          & 71.55    & 75.80          & 77.13          \\ \midrule
MT~\cite{Tarvainen2017MeanTA}    & 70.51     & 71.53   & 73.02  & 76.58          \\
CutMix~\cite{DBLP:conf/bmvc/FrenchLAMF20}     & 71.66       & 75.51     & 77.33    & 78.21    \\
CCT~\cite{Ouali2020SemiSupervisedSS}      & 71.86      & 73.68         & 76.51          & 77.40          \\
GCT~\cite{Ke2020GuidedCT}                 & 70.90    & 73.29     & 76.66     & 77.98          \\
CPS~\cite{Chen2021SemiSupervisedSS}      & 74.48    & 76.44    & 77.68    & 78.64          \\
U$^{2}$PL$^{\ast}$~\cite{Wang2022SemiSupervisedSS}    & 74.43       & 77.60      & 78.70          & 79.94          \\
PS-MT~\cite{Liu2022PerturbedAS}           & 75.50        & 78.20      & 78.72          & 79.76         \\
UniMatch~\cite{Yang2022RevisitingWC} &78.10 &78.40 &79.20 & - \\ 
CCVC~\cite{Wang2023ConflictBasedCC} &76.80  &\textbf{79.40}  &79.60 & - \\
AugSeg~\cite{Zhao2022AugmentationMA} &77.01 &77.31 &78.82   &- \\
\midrule 
\textbf{Ours}   &\textbf{78.32}     &78.88   &\textbf{79.83}  &\textbf{80.90}\\
\bottomrule
\end{tabular}
}
\begin{minipage}{8cm} 
\vspace{2mm}
\footnotesize \quad
${\ast}$ Results are reproduced on CPS~\cite{Chen2021SemiSupervisedSS} splits. 
\end{minipage}
\label{tab:pascal blended}
\end{table}

\begin{table}[t]
\centering
\caption{Comparison with state-of-the-art methods on Cityscapes validation set with mIoU results (\%) $\uparrow$. Labeled images are sampled from Cityscapes \emph{train} set which contains 2975 samples.}
\resizebox{\columnwidth}{!}{
\begin{tabular}{@{}l|cccc@{}}
\toprule
Method     & 1/16 (186) & 1/8 (372)      & 1/4 (744)  & 1/2 (1488)     \\ \midrule
Supervised                                  & 65.74  & 72.53  & 74.43  & 77.83\\ \midrule
MT~\cite{Tarvainen2017MeanTA}               & 69.03  & 72.06  & 74.20  & 78.15          \\
CutMix~\cite{DBLP:conf/bmvc/FrenchLAMF20}   & 67.06  & 71.83  & 76.36  & 78.25          \\
CCT~\cite{Ouali2020SemiSupervisedSS}        & 69.32  & 74.12  & 75.99  & 78.10          \\
GCT~\cite{Ke2020GuidedCT}                   & 66.75  & 72.66  & 76.11  & 78.34    \\
CPS~\cite{Chen2021SemiSupervisedSS}         & 74.72  & 77.62  & 79.21  & 80.21       \\
U$^2$PL~\cite{Wang2022SemiSupervisedSS}     & 70.30  & 74.37  & 76.47  & 79.05          \\
PS-MT~\cite{Liu2022PerturbedAS}             & -      & 76.89  & 77.60  & 79.09          \\ 
PCR~\cite{Xu2022SemisupervisedSS}           &73.41   &76.31   &78.40   &79.11 \\
GTA-Seg~\cite{Jin2023SemiSupervisedSS}      &69.38   &72.02   &76.08   &-  \\ 
UniMatch~\cite{Yang2022RevisitingWC}        &76.60   &77.90   &79.20   &79.50 \\ 
AugSeg~\cite{Zhao2022AugmentationMA} &75.22 &77.82 &79.56 &80.43 \\
\midrule 
\textbf{Ours}   &\textbf{77.10}     &\textbf{78.19}   &\textbf{79.88}  &\textbf{80.82} \\
\bottomrule
\end{tabular}
}
\label{tab:Cityscapes}
\end{table}

\subsection{Ablation Studies}
In this section, we conduct a series of experiments to investigate the effectiveness of our proposed feature perturbation strategy. All the experiments are based on 1/4 (366) and 1/2 (732) data partitions in \emph{classic} Pascal VOC 2012 dataset.  

\begin{table}[]
\centering
\caption{Ablation study on main components of DDFP. $\mathbf{L}_{con}^{im}$: Self-training with image-level consistency regularization. $\mathcal{L}_{con}^{ft}$: Feature-level consistency regularization. Random: Random noise sampled from normal distribution as perturbations. DD: The proposed density-descending perturbations.}
\resizebox{\columnwidth}{!}{
\begin{tabular}{@{}c|cccc|cc@{}}
\toprule
\multicolumn{1}{l|}{} & \multicolumn{4}{c|}{DDFP Framework}                                                                    & \multicolumn{2}{c}{mIoU (\%)}               \\ \midrule
\multicolumn{1}{l|}{} & \multicolumn{1}{c|}{\multirow{2}{*}{Supervised}} & \multicolumn{1}{c|}{\multirow{2}{*}{$\mathcal{L}_{con}^{im}$}} & \multicolumn{1}{c|}{\multirow{2}{*}{\begin{tabular}[c]{@{}c@{}}$\mathcal{L}_{con}^{ft}$\\ (Random)\end{tabular}}} & \multirow{2}{*}{\begin{tabular}[c]{@{}c@{}}$\mathcal{L}_{con}^{ft}$\\ (DD)\end{tabular}} & \multirow{2}{*}{366} & \multirow{2}{*}{732} \\
\multicolumn{1}{l|}{} & \multicolumn{1}{c|}{}                            & \multicolumn{1}{c|}{}                                          & \multicolumn{1}{c|}{}                                                                                             &                                                                                          &                      &                      \\ \midrule
I                     & \multicolumn{1}{c|}{\checkmark}                  & \multicolumn{1}{c|}{}                                          & \multicolumn{1}{c|}{}                                                                                             &                                                                                          & 65.88                & 71.69                \\
II                    & \multicolumn{1}{c|}{\checkmark}                  & \multicolumn{1}{c|}{\checkmark}                                & \multicolumn{1}{c|}{}                                                                                             &                                                                                          & 75.44                & 77.56                \\
III                   & \multicolumn{1}{c|}{\checkmark}                  & \multicolumn{1}{c|}{\checkmark}                                & \multicolumn{1}{c|}{\checkmark}                                                                                   &                                                                                          & 76.26                & 78.51                \\
IV                    & \multicolumn{1}{c|}{\checkmark}                  & \multicolumn{1}{c|}{\checkmark}                                & \multicolumn{1}{c|}{}                                                                                             & \checkmark                                                                               & \textbf{79.51}       & \textbf{81.21}       \\ \bottomrule
\end{tabular}
}
\label{tab:main ablation}
\end{table}

% \usepackage{booktabs}
\begin{table}[]
\centering
\caption{Ablation study on the magnitude of perturbation vectors. The magnitude refers to $\epsilon$ in Eq.~\ref{eq:perturbation calculation}.}
\begin{tabular}{@{}c|ccccc@{}}
\toprule
Magnitude & 1     & 2     & 4     & 6     & 8     \\ \midrule
366       & 78.76 & 78.96 & \textbf{79.51} & 79.43 & 79.41 \\
732       & 81.03 & \textbf{81.23} & 81.21 & 81.07 & 80.79 \\ \bottomrule
\end{tabular}
\label{tab:perturbation magnitude}
\end{table}

% \usepackage{booktabs}
\begin{table}[]
\centering
\caption{Ablation study on the impact of loss weight $\lambda_{ft}$ in Eq.~\ref{eq:overall optimization}.}
\begin{tabular}{@{}c|ccccc@{}}
\toprule
 $\lambda_{ft}$ & 0.2 & 0.5   & 1     & 1.5   & 2     \\ \midrule
366     & 78.89   & \textbf{79.51} & 79.28 & 79.40     & 77.78     \\
732     & 80.87   & \textbf{81.21} & 81.13 & 80.56 & 80.22 \\ \bottomrule
\end{tabular}
\label{tab:loss weight}
\end{table}

\begin{table}[]
\centering
\caption{Comparison with other feature-level perturbations. All perturbations are implemented on the same baseline framework.}
\begin{tabular}{@{}l|cc@{}}
\toprule
Perturbation Type  & 366   & 732   \\ \midrule
Uniform Noise~\cite{Ouali2020SemiSupervisedSS}      & 75.69 & 78.52 \\
Channel Dropout~\cite{Yang2022RevisitingWC}    & 77.76 & 80.38 \\
VAT~\cite{Liu2022PerturbedAS}                & 76.87 & 78.38 \\
Density-descending & \textbf{79.51} & \textbf{81.21} \\ \bottomrule
\end{tabular}
\label{tab: perturbation type}
\end{table}

\vspace{2mm}
\noindent
\textbf{Effectiveness of the density-descending perturbation.} In Tab.~\ref{tab:main ablation}, we ablate the framework to manifest the effectiveness of DDFP. We set the model with image-level consistency regularization in Experiment II as the main baseline. We first introduce noise sampled from normal distribution to see if randomly perturbed features can enhance model performance. The noise vectors are normalized and then applied on features. By comparing results in Experiment III and II, random perturbations improve baseline by +0.82\% and +0.95\% on 366 and 732 settings, respectively. Then, in Experiment IV, we inject density-descending perturbations under the same magnitude to replace random noise, which significantly boosts the model performance by +4.07\% and +3.65\% on two splits compared with baseline. This indicates that most performance gain in DDFP is brought by our density-descending design.  

\vspace{2mm}
\noindent
\textbf{Impact of perturbation step size $\epsilon$.} We also investigate the impact of perturbation step size $\epsilon$ in Eq.~\ref{eq:perturbation calculation}, which indicates the magnitude of a normalized vector. In Tab.~\ref{tab:perturbation magnitude}, we compare the performance under different step sizes. We found that the model performance improves when step size increases from 1, and the optimal step size slightly varies on different data splits. We choose step size of 4 as the default setting where overall best performance can be achieved. Overly large step size can hurt the performance since aggressive exploring towards low density regions might end up with out-of-distribution samples.   

\vspace{2mm}
\noindent
\textbf{Impact of loss weight $\lambda_{ft}$.} We also examine the model performance under different $\lambda_{ft}$ in Eq.~\ref{eq:overall optimization}. As shown in Tab.~\ref{tab:loss weight}, the optimal weight for our feature-level consistency loss is around 0.5, which is the default setting in our experiments. Further increasing the contribution of the loss can lead to degenerated performance. Our guess is the image-level consistency serves as the foundation for the proposed perturbation strategy to function and the overly aggressive optimization of feature-level consistency might interfere the consistency learning at image-level. 

\vspace{2mm}
\noindent
\textbf{Comparison with other feature perturbations.} To further validate the effectiveness of DDFP, we compare it with different types of feature-level perturbations. As shown in Tab.~\ref{tab: perturbation type}, Uniform Noise is the perturbation sampled from a uniform distribution as in CCT~\cite{Ouali2020SemiSupervisedSS}. Channel Dropout refers to randomly zeroing out half of the feature channels following UniMatch~\cite{Yang2022RevisitingWC}. VAT indicates the perturbation in virtual adversarial training~\cite{Miyato2017VirtualAT} which is introduced in semi-supervised semantic segmentation by PS-MT~\cite{Liu2022PerturbedAS}. All the perturbations are implemented on the same baseline framework which is self-training with single-stream weak-to-strong image-level consistency regularization. In Tab.~\ref{tab: perturbation type}, the quantitative results show the superiority of our density-descending perturbation. Specifically, our DDFP significantly outperforms the second best Channel Dropout strategy by +1.75\% and +0.83\% on 366 and 732 splits.      

% Unlike PS-MT with dual teachers, we only adopt a single teacher model for perturbation generation.

\begin{figure}
    \centering
    \includegraphics[width=\columnwidth]{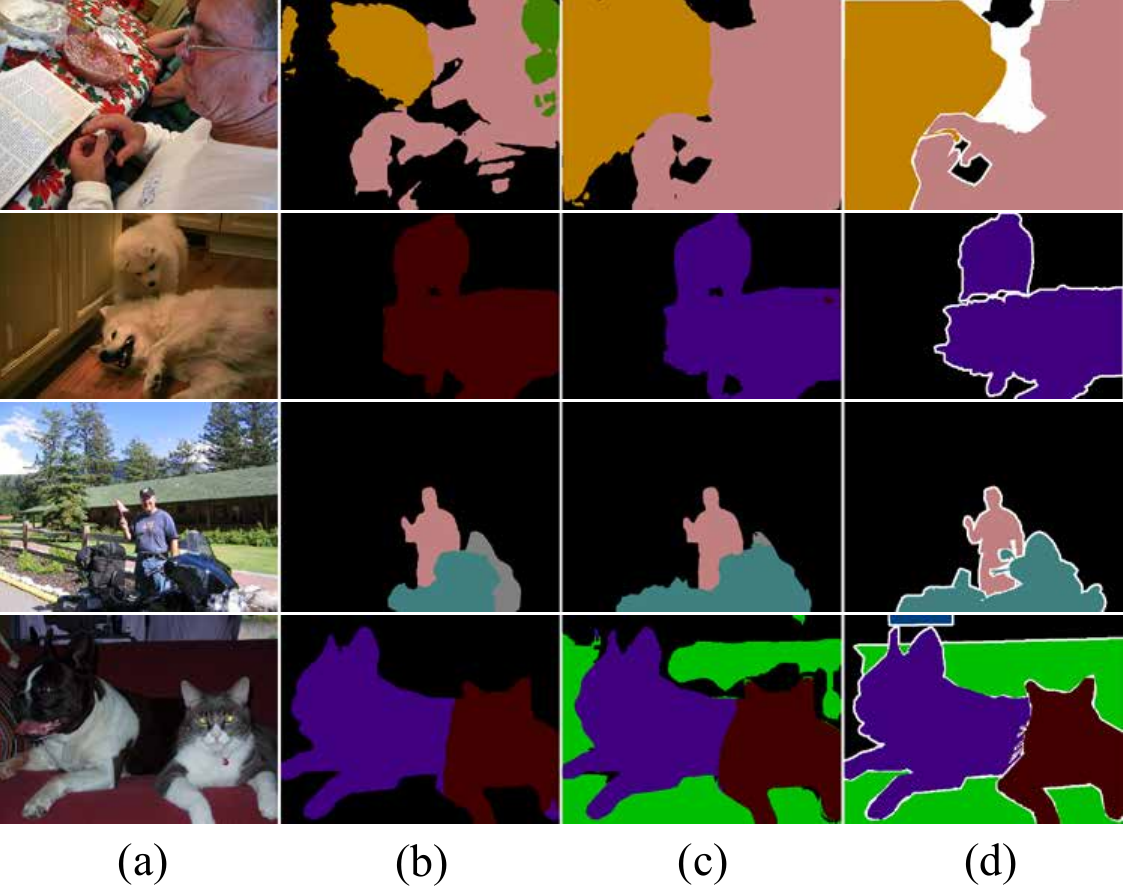}
    \caption{Qualitative results on Pascal VOC 2012 dataset. Models are trained under \emph{classic} 732 partitions. (a) Input images. (b) Results from baseline model with image-level consistency regularization only. (c) Results by our DDFP. (d) Ground truth.}
    \label{fig:qualitative}
\end{figure}

\vspace{2mm}
\noindent
\textbf{Qualitative Results.} We show the qualitative results on Pascal VOC dataset in Fig.~\ref{fig:qualitative}. By observation, baseline method only with image-level consistency regularization performs poorly to capture the complete outline of the object as shown in the first row in Fig.~\ref{fig:qualitative} (b). Also, in certain context, it has difficulty distinguishing visually similar classes like cat and dog as shown in the second row of Fig.~\ref{fig:qualitative} (b). Equipped with DDFP, the model predicted more accurately on shapes and classes, which is shown in Fig.~\ref{fig:qualitative} (c).   

\section{Conclusion}
In this work, we propose a novel feature-level consistency regularization strategy name Density-Descending Feature Perturbation (DDFP) for semi-supervised semantic segmentation. The aim of DDFP is to create perturbed features in low density regions in feature space, to force decision boundary to explore less dense regions thus enhancing model generalization. Density estimation is the heart of our method, which is achieved by our proposed lightweight density estimator based on normalizing flow. Extensive experiments under various data settings have shown that our DDFP can effectively boost model performance and outperform other types of feature-level perturbation designs.

\clearpage
{
    \small
    \bibliographystyle{ieeenat_fullname}
    \bibliography{main}

\begin{thebibliography}{59}
\providecommand{\natexlab}[1]{#1}
\providecommand{\url}[1]{\texttt{#1}}
\expandafter\ifx\csname urlstyle\endcsname\relax
  \providecommand{\doi}[1]{doi: #1}\else
  \providecommand{\doi}{doi: \begingroup \urlstyle{rm}\Url}\fi

\bibitem[Alonso et~al.(2021)Alonso, Sabater, Ferstl, Montesano, and Murillo]{Alonso2021SemiSupervisedSS}
I{\~n}igo Alonso, Alberto Sabater, David Ferstl, Luis Montesano, and Ana~Cristina Murillo.
\newblock Semi-supervised semantic segmentation with pixel-level contrastive learning from a class-wise memory bank.
\newblock In \emph{ICCV}, 2021.

\bibitem[Ardizzone et~al.(2022)Ardizzone, Bungert, Draxler, Köthe, Kruse, Schmier, and Sorrenson]{freia}
Lynton Ardizzone, Till Bungert, Felix Draxler, Ullrich Köthe, Jakob Kruse, Robert Schmier, and Peter Sorrenson.
\newblock Framework for easily invertible architectures (freia), 2022.

\bibitem[Berthelot et~al.(2019)Berthelot, Carlini, Goodfellow, Papernot, Oliver, and Raffel]{Berthelot2019MixMatchAH}
David Berthelot, Nicholas Carlini, Ian~J. Goodfellow, Nicolas Papernot, Avital Oliver, and Colin Raffel.
\newblock Mixmatch: A holistic approach to semi-supervised learning.
\newblock In \emph{NeurIPS}, 2019.

\bibitem[Berthelot et~al.(2020)Berthelot, Carlini, Cubuk, Kurakin, Sohn, Zhang, and Raffel]{Berthelot2020ReMixMatchSL}
David Berthelot, Nicholas Carlini, Ekin~Dogus Cubuk, Alexey Kurakin, Kihyuk Sohn, Han Zhang, and Colin Raffel.
\newblock Remixmatch: Semi-supervised learning with distribution matching and augmentation anchoring.
\newblock In \emph{ICLR}, 2020.

\bibitem[Chapelle and Zien(2005)]{Chapelle2005SemiSupervisedCB}
Olivier Chapelle and Alexander Zien.
\newblock Semi-supervised classification by low density separation.
\newblock In \emph{International Conference on Artificial Intelligence and Statistics}, 2005.

\bibitem[Chen et~al.(2023)Chen, Tao, Fan, Wang, Wang, Schiele, Xie, Raj, and Savvides]{Chen2023SoftMatchAT}
Hao Chen, Ran Tao, Yue Fan, Yidong Wang, Jindong Wang, Bernt Schiele, Xingxu Xie, Bhiksha Raj, and Marios Savvides.
\newblock Softmatch: Addressing the quantity-quality trade-off in semi-supervised learning.
\newblock In \emph{ICLR}, 2023.

\bibitem[Chen et~al.(2018{\natexlab{a}})Chen, Papandreou, Kokkinos, Murphy, and Yuille]{deeplab}
Liang-Chieh Chen, George Papandreou, Iasonas Kokkinos, Kevin Murphy, and Alan~L. Yuille.
\newblock Deeplab: Semantic image segmentation with deep convolutional nets, atrous convolution, and fully connected crfs.
\newblock \emph{PAMI}, 40\penalty0 (4):\penalty0 834--848, 2018{\natexlab{a}}.

\bibitem[Chen et~al.(2018{\natexlab{b}})Chen, Zhu, Papandreou, Schroff, and Adam]{Chen2018EncoderDecoderWA}
Liang-Chieh Chen, Yukun Zhu, George Papandreou, Florian Schroff, and Hartwig Adam.
\newblock Encoder-decoder with atrous separable convolution for semantic image segmentation.
\newblock In \emph{ECCV}, 2018{\natexlab{b}}.

\bibitem[Chen et~al.(2021)Chen, Yuan, Zeng, and Wang]{Chen2021SemiSupervisedSS}
Xiaokang Chen, Yuhui Yuan, Gang Zeng, and Jingdong Wang.
\newblock Semi-supervised semantic segmentation with cross pseudo supervision.
\newblock In \emph{CVPR}, 2021.

\bibitem[Cordts et~al.(2016)Cordts, Omran, Ramos, Rehfeld, Enzweiler, Benenson, Franke, Roth, and Schiele]{cityscapes}
Marius Cordts, Mohamed Omran, Sebastian Ramos, Timo Rehfeld, Markus Enzweiler, Rodrigo Benenson, Uwe Franke, Stefan Roth, and Bernt Schiele.
\newblock The cityscapes dataset for semantic urban scene understanding.
\newblock In \emph{CVPR}, 2016.

\bibitem[Deng et~al.(2009)Deng, Dong, Socher, Li, Li, and Fei-Fei]{Deng2009ImageNetAL}
Jia Deng, Wei Dong, Richard Socher, Li-Jia Li, K. Li, and Li Fei-Fei.
\newblock Imagenet: A large-scale hierarchical image database.
\newblock In \emph{CVPR}, 2009.

\bibitem[Dinh et~al.(2017)Dinh, Sohl-Dickstein, and Bengio]{Dinh2016DensityEU}
Laurent Dinh, Jascha~Narain Sohl-Dickstein, and Samy Bengio.
\newblock Density estimation using real nvp.
\newblock \emph{ICLR}, 2017.

\bibitem[Everingham et~al.(2009)Everingham, Gool, Williams, Winn, and Zisserman]{pascal}
Mark Everingham, Luc~Van Gool, Christopher K.~I. Williams, John~M. Winn, and Andrew Zisserman.
\newblock The pascal visual object classes (voc) challenge.
\newblock \emph{IJCV}, 88\penalty0 (2):\penalty0 303--338, 2009.

\bibitem[Feng et~al.(2020)Feng, Zhou, Gu, Tan, Cheng, Lu, Shi, and Ma]{Feng2020DMTDM}
Zhengyang Feng, Qianyu Zhou, Qiqi Gu, Xin Tan, Guangliang Cheng, Xuequan Lu, Jianping Shi, and Lizhuang Ma.
\newblock Dmt: Dynamic mutual training for semi-supervised learning.
\newblock \emph{PR}, 130:\penalty0 108777, 2020.

\bibitem[French et~al.(2020)French, Laine, Aila, Mackiewicz, and Finlayson]{DBLP:conf/bmvc/FrenchLAMF20}
Geoffrey French, Samuli Laine, Timo Aila, Michal Mackiewicz, and Graham~D. Finlayson.
\newblock Semi-supervised semantic segmentation needs strong, varied perturbations.
\newblock In \emph{BMVC}, 2020.

\bibitem[Guan et~al.(2022)Guan, Huang, Xiao, and Lu]{Guan2022UnbiasedSR}
Dayan Guan, Jiaxing Huang, Aoran Xiao, and Shijian Lu.
\newblock Unbiased subclass regularization for semi-supervised semantic segmentation.
\newblock In \emph{CVPR}, 2022.

\bibitem[Hariharan et~al.(2011)Hariharan, Arbel{\'a}ez, Bourdev, Maji, and Malik]{sbd}
Bharath Hariharan, Pablo Arbel{\'a}ez, Lubomir~D. Bourdev, Subhransu Maji, and Jitendra Malik.
\newblock Semantic contours from inverse detectors.
\newblock In \emph{ICCV}, 2011.

\bibitem[He et~al.(2016)He, Zhang, Ren, and Sun]{He2016DeepRL}
Kaiming He, X. Zhang, Shaoqing Ren, and Jian Sun.
\newblock Deep residual learning for image recognition.
\newblock In \emph{CVPR}, 2016.

\bibitem[He et~al.(2021)He, Yang, and Qi]{He2021RedistributingBP}
Ruifei He, Jihan Yang, and Xiaojuan Qi.
\newblock Re-distributing biased pseudo labels for semi-supervised semantic segmentation: A baseline investigation.
\newblock In \emph{ICCV}, 2021.

\bibitem[Hu et~al.(2021)Hu, Wei, Hu, Ye, Cui, and Wang]{Hu2021SemiSupervisedSS}
Hanzhe Hu, Fangyun Wei, Han Hu, Qiwei Ye, Jinshi Cui, and Liwei Wang.
\newblock Semi-supervised semantic segmentation via adaptive equalization learning.
\newblock In \emph{NeurIPS}, 2021.

\bibitem[Hung et~al.(2018)Hung, Tsai, Liou, Lin, and Yang]{Hung2018AdversarialLF}
Wei-Chih Hung, Yi-Hsuan Tsai, Yan-Ting Liou, Yen-Yu Lin, and Ming-Hsuan Yang.
\newblock Adversarial learning for semi-supervised semantic segmentation.
\newblock In \emph{BMVC}, 2018.

\bibitem[Izmailov et~al.(2019)Izmailov, Kirichenko, Finzi, and Wilson]{Izmailov2019SemiSupervisedLW}
Pavel Izmailov, P. Kirichenko, Marc Finzi, and Andrew~Gordon Wilson.
\newblock Semi-supervised learning with normalizing flows.
\newblock In \emph{ICML}, 2019.

\bibitem[Jeong et~al.(2019)Jeong, Lee, Kim, and Kwak]{Jeong2019ConsistencybasedSL}
Jisoo Jeong, Seungeui Lee, Jeesoo Kim, and Nojun Kwak.
\newblock Consistency-based semi-supervised learning for object detection.
\newblock In \emph{NeurIPS}, 2019.

\bibitem[Jin et~al.(2022)Jin, Wang, and Lin]{Jin2023SemiSupervisedSS}
Ying Jin, Jiaqi Wang, and Dahua Lin.
\newblock Semi-supervised semantic segmentation via gentle teaching assistant.
\newblock In \emph{NeurIPS}, 2022.

\bibitem[Ke et~al.(2020)Ke, Qiu, Li, Yan, and Lau]{Ke2020GuidedCT}
Zhanghan Ke, Di Qiu, Kaican Li, Qiong Yan, and Rynson W.~H. Lau.
\newblock Guided collaborative training for pixel-wise semi-supervised learning.
\newblock In \emph{ECCV}, 2020.

\bibitem[Kingma and Dhariwal(2018)]{Kingma2018GlowGF}
Diederik~P. Kingma and Prafulla Dhariwal.
\newblock Glow: Generative flow with invertible 1x1 convolutions.
\newblock In \emph{NeurIPS}, 2018.

\bibitem[Lai et~al.(2021)Lai, Tian, Jiang, Liu, Zhao, Wang, and Jia]{Lai2021SemisupervisedSS}
Xin Lai, Zhuotao Tian, Li Jiang, Shu Liu, Hengshuang Zhao, Liwei Wang, and Jiaya Jia.
\newblock Semi-supervised semantic segmentation with directional context-aware consistency.
\newblock In \emph{CVPR}, 2021.

\bibitem[Laine and Aila(2017)]{Laine2017TemporalEF}
Samuli Laine and Timo Aila.
\newblock Temporal ensembling for semi-supervised learning.
\newblock In \emph{ICLR}, 2017.

\bibitem[Liu et~al.(2022{\natexlab{a}})Liu, Zhang, Bai, Zhang, and Zhao]{9656684}
Man Liu, Chunjie Zhang, Huihui Bai, Riquan Zhang, and Yao Zhao.
\newblock Cross-part learning for fine-grained image classification.
\newblock \emph{TIP}, 31:\penalty0 748--758, 2022{\natexlab{a}}.

\bibitem[Liu et~al.(2023)Liu, Li, Zhang, Wei, Bai, and Zhao]{Liu2023ProgressiveSM}
Man Liu, Feng Li, Chunjie Zhang, Yunchao Wei, Huihui Bai, and Yao Zhao.
\newblock Progressive semantic-visual mutual adaption for generalized zero-shot learning.
\newblock In \emph{CVPR}, 2023.

\bibitem[Liu et~al.(2022{\natexlab{b}})Liu, Zhi, Johns, and Davison]{liu2022reco}
Shikun Liu, Shuaifeng Zhi, Edward Johns, and Andrew~J Davison.
\newblock Bootstrapping semantic segmentation with regional contrast.
\newblock In \emph{ICLR}, 2022{\natexlab{b}}.

\bibitem[Liu et~al.(2022{\natexlab{c}})Liu, Tian, Chen, Liu, Belagiannis, and Carneiro]{Liu2022PerturbedAS}
Yuyuan Liu, Yu Tian, Yuanhong Chen, Fengbei Liu, Vasileios Belagiannis, and G. Carneiro.
\newblock Perturbed and strict mean teachers for semi-supervised semantic segmentation.
\newblock In \emph{CVPR}, 2022{\natexlab{c}}.

\bibitem[Long et~al.(2015)Long, Shelhamer, and Darrell]{fcn}
Jonathan Long, Evan Shelhamer, and Trevor Darrell.
\newblock Fully convolutional networks for semantic segmentation.
\newblock In \emph{CVPR}, 2015.

\bibitem[Mittal et~al.(2021)Mittal, Tatarchenko, and Brox]{Mittal2021SemiSupervisedSS}
Sudhanshu Mittal, Maxim Tatarchenko, and Thomas Brox.
\newblock Semi-supervised semantic segmentation with high- and low-level consistency.
\newblock \emph{PAMI}, 43\penalty0 (4):\penalty0 1369--1379, 2021.

\bibitem[Miyato et~al.(2019)Miyato, Maeda, Koyama, and Ishii]{Miyato2017VirtualAT}
Takeru Miyato, Shin{-}ichi Maeda, Masanori Koyama, and Shin Ishii.
\newblock Virtual adversarial training: {A} regularization method for supervised and semi-supervised learning.
\newblock \emph{PAMI}, 41\penalty0 (8):\penalty0 1979--1993, 2019.

\bibitem[Ouali et~al.(2020)Ouali, Hudelot, and Tami]{Ouali2020SemiSupervisedSS}
Yassine Ouali, C{\'e}line Hudelot, and Myriam Tami.
\newblock Semi-supervised semantic segmentation with cross-consistency training.
\newblock In \emph{CVPR}, 2020.

\bibitem[Pham et~al.(2021)Pham, Xie, Dai, and Le]{Pham2020MetaPL}
Hieu Pham, Qizhe Xie, Zihang Dai, and Quoc~V. Le.
\newblock Meta pseudo labels.
\newblock In \emph{CVPR}, 2021.

\bibitem[Rosenberg et~al.(2005)Rosenberg, Hebert, and Schneiderman]{Rosenberg2005SemiSupervisedSO}
Chuck Rosenberg, Martial Hebert, and Henry Schneiderman.
\newblock Semi-supervised self-training of object detection models.
\newblock In \emph{WACV}, 2005.

\bibitem[Sajjadi et~al.(2016)Sajjadi, Javanmardi, and Tasdizen]{Sajjadi2016RegularizationWS}
Mehdi S.~M. Sajjadi, Mehran Javanmardi, and Tolga Tasdizen.
\newblock Regularization with stochastic transformations and perturbations for deep semi-supervised learning.
\newblock In \emph{NeurIPS}, 2016.

\bibitem[Sohn et~al.(2020)Sohn, Berthelot, Li, Zhang, Carlini, Cubuk, Kurakin, Zhang, and Raffel]{Sohn2020FixMatchSS}
Kihyuk Sohn, David Berthelot, Chun-Liang Li, Zizhao Zhang, Nicholas Carlini, Ekin~Dogus Cubuk, Alexey Kurakin, Han Zhang, and Colin Raffel.
\newblock Fixmatch: Simplifying semi-supervised learning with consistency and confidence.
\newblock In \emph{NeurIPS}, 2020.

\bibitem[Souly et~al.(2017)Souly, Spampinato, and Shah]{Souly2017SemiSS}
Nasim Souly, Concetto Spampinato, and Mubarak Shah.
\newblock Semi supervised semantic segmentation using generative adversarial network.
\newblock In \emph{ICCV}, 2017.

\bibitem[Tarvainen and Valpola(2017)]{Tarvainen2017MeanTA}
Antti Tarvainen and Harri Valpola.
\newblock Mean teachers are better role models: Weight-averaged consistency targets improve semi-supervised deep learning results.
\newblock In \emph{NeurIPS}, 2017.

\bibitem[Wang et~al.(2023{\natexlab{a}})Wang, Zhang, Yu, and Xiao]{Wang2023HuntingSD}
Xiaoyang Wang, Bingfeng Zhang, Limin Yu, and Jimin Xiao.
\newblock Hunting sparsity: Density-guided contrastive learning for semi-supervised semantic segmentation.
\newblock In \emph{CVPR}, 2023{\natexlab{a}}.

\bibitem[Wang et~al.(2022)Wang, Wang, Shen, Fei, Li, Jin, Wu, Zhao, and Le]{Wang2022SemiSupervisedSS}
Yuchao Wang, Haochen Wang, Yujun Shen, Jingjing Fei, Wei Li, Guoqiang Jin, Liwei Wu, Rui Zhao, and Xinyi Le.
\newblock Semi-supervised semantic segmentation using unreliable pseudo-labels.
\newblock In \emph{CVPR}, 2022.

\bibitem[Wang et~al.(2023{\natexlab{b}})Wang, Chen, Heng, Hou, Savvides, Shinozaki, Raj, Wu, and Wang]{Wang2022FreeMatchST}
Yidong Wang, Hao Chen, Qiang Heng, Wenxin Hou, Marios Savvides, Takahiro Shinozaki, Bhiksha Raj, Zhen Wu, and Jindong Wang.
\newblock Freematch: Self-adaptive thresholding for semi-supervised learning.
\newblock In \emph{ICLR}, 2023{\natexlab{b}}.

\bibitem[Wang et~al.(2023{\natexlab{c}})Wang, Zhao, Zhou, Xu, Xing, and Kong]{Wang2023ConflictBasedCC}
Zicheng Wang, Zhen Zhao, Luping Zhou, Dong Xu, Xiaoxia Xing, and Xiangyu Kong.
\newblock Conflict-based cross-view consistency for semi-supervised semantic segmentation.
\newblock In \emph{CVPR}, 2023{\natexlab{c}}.

\bibitem[Xie et~al.(2020)Xie, Hovy, Luong, and Le]{Xie2020SelfTrainingWN}
Qizhe Xie, Eduard~H. Hovy, Minh-Thang Luong, and Quoc~V. Le.
\newblock Self-training with noisy student improves imagenet classification.
\newblock In \emph{CVPR}, 2020.

\bibitem[Xu et~al.(2022)Xu, Liu, Bian, and Yang]{Xu2022SemisupervisedSS}
Hai-Ming Xu, Lingqiao Liu, Qiuchen Bian, and Zhengeng Yang.
\newblock Semi-supervised semantic segmentation with prototype-based consistency regularization.
\newblock In \emph{NeurIPS}, 2022.

\bibitem[Yang et~al.(2022)Yang, Zhuo, Qi, Shi, and Gao]{Yang2022STMS}
Lihe Yang, Wei Zhuo, Lei Qi, Yinghuan Shi, and Yang Gao.
\newblock St++: Make self-training work better for semi-supervised semantic segmentation.
\newblock In \emph{CVPR}, 2022.

\bibitem[Yang et~al.(2023)Yang, Qi, Feng, Zhang, and Shi]{Yang2022RevisitingWC}
Lihe Yang, Lei Qi, Litong Feng, Wayne Zhang, and Yinghuan Shi.
\newblock Revisiting weak-to-strong consistency in semi-supervised semantic segmentation.
\newblock In \emph{CVPR}, 2023.

\bibitem[Yuan et~al.(2021)Yuan, Liu, Shen, Wang, and Li]{Yuan2021ASB}
Jianlong Yuan, Yifan Liu, Chunhua Shen, Zhibin Wang, and Hao Li.
\newblock A simple baseline for semi-supervised semantic segmentation with strong data augmentation.
\newblock In \emph{ICCV}, 2021.

\bibitem[Zhang et~al.(2021)Zhang, Wang, Hou, Wu, Wang, Okumura, and Shinozaki]{Zhang2021FlexMatchBS}
Bowen Zhang, Yidong Wang, Wenxin Hou, Hao Wu, Jindong Wang, Manabu Okumura, and Takahiro Shinozaki.
\newblock Flexmatch: Boosting semi-supervised learning with curriculum pseudo labeling.
\newblock In \emph{NeurIPS}, 2021.

\bibitem[Zhao et~al.(2017)Zhao, Shi, Qi, Wang, and Jia]{Zhao2017PyramidSP}
Hengshuang Zhao, Jianping Shi, Xiaojuan Qi, Xiaogang Wang, and Jiaya Jia.
\newblock Pyramid scene parsing network.
\newblock In \emph{CVPR}, 2017.

\bibitem[Zhao et~al.(2021)Zhao, Vemulapalli, Mansfield, Gong, Green, Shapira, and Wu]{Zhao2021ContrastiveLF}
Xiangyu Zhao, Raviteja Vemulapalli, P.~A. Mansfield, Boqing Gong, Bradley Green, Lior Shapira, and Ying Wu.
\newblock Contrastive learning for label efficient semantic segmentation.
\newblock In \emph{ICCV}, 2021.

\bibitem[Zhao et~al.(2024)Zhao, Tang, Wang, and Xiao]{Zhao2024SFCSF}
Xinqiao Zhao, Feilong Tang, Xiaoyang Wang, and Jimin Xiao.
\newblock Sfc: Shared feature calibration in weakly supervised semantic segmentation.
\newblock In \emph{AAAI}, 2024.

\bibitem[Zhao et~al.(2023)Zhao, Yang, Long, Pi, Zhou, and Wang]{Zhao2022AugmentationMA}
Zhen Zhao, Lihe Yang, Sifan Long, Jimin Pi, Luping Zhou, and Jingdong Wang.
\newblock Augmentation matters: A simple-yet-effective approach to semi-supervised semantic segmentation.
\newblock In \emph{CVPR}, 2023.

\bibitem[Zhong et~al.(2021)Zhong, Yuan, Wu, Yuan, Peng, and Wang]{Zhong2021PixelCS}
Yuanyi Zhong, Bodi Yuan, Hong Wu, Zhiqiang Yuan, Jian Peng, and Yu-Xiong Wang.
\newblock Pixel contrastive-consistent semi-supervised semantic segmentation.
\newblock In \emph{ICCV}, 2021.

\bibitem[Zhou et~al.(2021)Zhou, Xu, Zhang, Gao, and Heng]{Zhou2021C3SemiSegCS}
Yanning Zhou, Hang Xu, Wei Zhang, Bin-Bin Gao, and Pheng-Ann Heng.
\newblock C3-semiseg: Contrastive semi-supervised segmentation via cross-set learning and dynamic class-balancing.
\newblock In \emph{ICCV}, 2021.

\bibitem[Zoph et~al.(2020)Zoph, Ghiasi, Lin, Cui, Liu, Cubuk, and Le]{Zoph2020RethinkingPA}
Barret Zoph, Golnaz Ghiasi, Tsung-Yi Lin, Yin Cui, Hanxiao Liu, Ekin~Dogus Cubuk, and Quoc~V. Le.
\newblock Rethinking pre-training and self-training.
\newblock In \emph{NeurIPS}, 2020.

\end{thebibliography}
}


\end{document}